\documentclass{article}
\usepackage{spconf,amsmath,graphicx}
\usepackage{multirow}
\usepackage{glossaries}
\usepackage{amssymb}
\usepackage{cite}
\usepackage{tikz, svg}
\usepackage{subfig}
\usepackage{booktabs}
\usepackage{url}
\usepackage[export]{adjustbox}
\usepackage{placeins}
\usepackage{diagbox}
\usepackage[
detect-weight,
exponent-product=\cdot,
locale = US,
group-separator=.,
]{siunitx}

\tikzstyle{none}=[inner sep=0pt]
\tikzstyle{layer}=[fill={rgb,255: red,174; green,199; blue,232}, draw=black, shape=rectangle, minimum width=0.25cm, minimum height=1.5cm]
\tikzstyle{decoder}=[fill={rgb,255: red,255; green,187; blue,120}, draw=black, shape=rectangle, minimum width=0.25cm, minimum height=1.5cm]
\tikzstyle{linear}=[fill={rgb,255: red,152; green,223; blue,138}, draw=black, shape=rectangle, minimum width=0.25cm, minimum height=1.5cm]
\tikzstyle{add}=[fill=none, draw=black, shape=circle]
\tikzstyle{emb}=[fill={rgb,255: red,197; green,176; blue,213}, draw=black, shape=rectangle, minimum width=0.25cm, minimum height=0.6cm]
\tikzstyle{onehot}=[fill=white, draw=black, shape=rectangle, minimum width=0.25cm, minimum height=0.6cm]
\tikzstyle{posenc}=[fill={rgb,255: red,255; green,152; blue,150}, draw=black, shape=rectangle, minimum width=0.25cm, minimum height=0.6cm]
\tikzstyle{borderless}=[fill=white, draw=white, shape=circle]
\tikzstyle{op}=[fill={rgb,255: red,224; green,224; blue,224}, draw=black, shape=circle, minimum size=0.5cm, scale=0.7]
\tikzstyle{arrow}=[draw=black, ->, line width=0.4mm]
\tikzstyle{new edge style 0}=[{|-|},  line width=0.4mm]

\pgfdeclarelayer{nodelayer}
\pgfdeclarelayer{edgelayer}
\pgfsetlayers{edgelayer,nodelayer,main}

\newacronym{lp}{LP}{license plate}
\newacronym{lpr}{LPR}{license plate recognition}
\newacronym{alpr}{ALPR}{automatic license plate recognition}
\newacronym{flpr}{FLPR}{forensic license plate recognition}
\newacronym{cnn}{CNN}{convolutional neural network}
\newacronym{crnn}{CRNN}{convolutional recurrent neural network}
\newacronym{rnn}{RNN}{recurrent neural network}
\newacronym{nn}{NN}{neural network}
\newacronym{qf}{QF}{quality factor}
\newacronym{qp}{QP}{quantization parameter}
\newacronym{seq2seq}{Seq2Seq}{sequence-to-sequence}

\def\acclp{{ acc\textsubscript{lp} }}
\DeclareSIUnit{\pp}{\textup{p.p.}}

\title{Forensic License Plate Recognition with Compression-Informed Transformers}

\name{Denise~Moussa$^{1,2}$, Anatol Maier$^{2}$, Andreas~Spruck$^{3}$, Jürgen~Seiler$^3$, Christian~Riess$^{2,}$\footnotemark[2]\sthanks{We gratefully acknowledge support by the German Federal Ministry of Education and Research (BMBF) under Grant No. 13N15319}}
\address{	$^1$Federal Criminal Police Office (BKA), Germany \\
	$^2$IT Security Infrastructures Lab, Computer Science, Univ. of Erlangen-N\"urnberg \\ $^3$Multimedia Communications and Signal Processing, Electrical Engineering, Univ. of Erlangen-Nürnberg}
\begin{document}
	
\ninept
\maketitle
\begin{abstract}
\Gls{flpr} remains an open challenge in legal contexts such as criminal
investigations, where unreadable \glspl{lp}  need to be deciphered from  highly compressed and/or low resolution footage, e.g., from surveillance cameras. In this work, we propose a side-informed Transformer architecture that embeds knowledge on the input compression level to improve recognition under strong compression.
We show the effectiveness of Transformers for \gls{lpr} on a low-quality real-world dataset. We also provide a synthetic dataset that includes strongly degraded, illegible LP images and analyze the impact of knowledge embedding on it. The network outperforms existing \gls{flpr} methods and standard state-of-the art image recognition models while requiring less parameters. For the severest degraded images, we can improve recognition by up to $8.9$ percent points.\footnote{The source code and datasets are available at \url{https://www.cs1.tf.fau.de/research/multimedia-security/}}

\end{abstract}
\section{Introduction}
\label{sec:intro}
\Gls{lp} detection and recognition is an active field of deep learning research. \gls{flpr} in particular aims at supporting criminal investigations by recognizing visually unreadable \gls{lp} characters from severely degraded footage. 
Hereby, the data mostly stems from uncontrolled sources, often from surveillance cameras of third parties.
Unfortunately, many commercial security cameras are
budget-constrained systems using strong compression and low resolution to achieve high memory and cost efficiency~\cite{kaiser2021}. Besides, these techniques amplify quality loss introduced by environmental factors like a vehicle's velocity and distance to the camera or weather and lighting conditions. Despite all these complicating factors for \gls{lpr}, pioneering works show that \glspl{nn} are still able to extract some \gls{lp} information from synthetic very low-quality images where human analysts and traditional forensic image enhancement fail~\cite{agarwal2017deciphering, lorch2019forensic, kaiser2021, rossi2021denoiser, moussa2021sequence}.

In this work, we aim at pushing the boundary for \gls{flpr} even further. Since to our knowledge there is no real world \gls{lp} dataset featuring visually unreadable \gls{lp} images, we generate SynthGLP, a synthetic dataset covering strong compression and low resolution, the arguably most challenging degradation factors for criminal investigations. We apply the JPEG compression~\cite{wallace1992jpeg} algorithm as it is the most widely spread lossy compression technique used by cameras and image editing tools and thus often appears in forensic image analysis~\cite{thai2016jpeg}. Regarding the architecture, we propose to use a Transformer network~\cite{vaswani2017attention} which includes prior knowledge about the compression strength to guide character recognition. It requires considerably less network parameters than the state-of-the-art, but outperforms existing works. As a simple scalar measure for compression strength we use the JPEG quality factor (QF). We demonstrate that our method is applicable for libjpeg QFs (coarsely) estimated from image data compressed with arbitrary quantization matrices. To the best of our knowledge, this is the first work that uses a Transformer on the task of (F)LPR. Hence, we demonstrate its applicability on a low-quality real world \gls{alpr} dataset. To summarize, the specific contributions of this work are:
\begin{itemize}
	\item We propose a parameter-efficient Transformer~\cite{vaswani2017attention} model for \gls{flpr} and evaluate it on real world data
	\item We show the added benefit of embedding compression knowledge at the example of JPEG compression
	\item We analyze performance on severely degraded, visually illegible synthetic \glspl{lp} in detail
\end{itemize}
The remainder of this work is structured in four parts. Section~\ref{sec:related_work} summarizes the related work, Sec.~\ref{sec:methods} describes our problem formulation, proposed network architecture and the used datasets. Section~\ref{sec:evaluation} presents our experiments and Sec.~\ref{sec:conclusion} concludes this work.
\begin{figure*}[t]
	\begin{center}
		\input{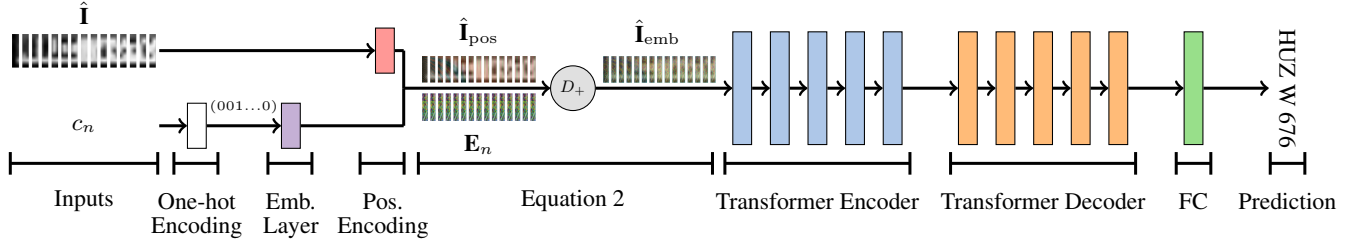}
	\end{center}
	\caption{The proposed Transformer architecture with knowledge embedding.}
	\label{fig:architecture}
\end{figure*}

\section{Related Work}
\label{sec:related_work}
We distinguish \gls{alpr} and \gls{flpr} systems for \gls{lp} identification. Many existing works address \gls{alpr} systems, which meet other requirements than \gls{flpr}. The main focus of \gls{alpr} systems is to automatically identify \gls{lp} strings from (at least partially) controlled acquisition systems of reasonably high quality. Applications include traffic monitoring, automatic toll collection or access control~\cite{sibgrapi_estendido}. The systems typically consist of an \gls{lp} detection step and an \gls{lpr} step. For \gls{lp} detection, commonly  YOLO-based
networks are used~\cite{laroca2018robust,silva2018license,zhang2020robust, sibgrapi_estendido}. \Gls{lp} recognition is generally implemented via \glspl{cnn}~\cite{laroca2018robust, silva2018license, sibgrapi_estendido} or \glspl{crnn}~\cite{shi2016end, shivakumara2018cnn, suvarnam2019combination, zhang2019license,zhang2020robust}. \gls{cnn} methods classify each character separately. \glspl{crnn} consider the \gls{lp} as a stream, which allows to flexibly process character sequences of varying lengths.

Several works recently paved the way for \gls{flpr}, which aims at deciphering unreadable \glspl{lp}. \v{S}pa{\v{n}}hel~\emph{et al.}~\cite{vspavnhel2017holistic} presented
a \gls{cnn} recognition method together with a low quality real world
\gls{lp} dataset of Czech \glspl{lp} called ReId. However, although ReId includes low-quality samples, they are not severely degraded, i.e., indecipherable.
Agarwal~\emph{et al.}~\cite{agarwal2017deciphering} are the first to 
show that \glspl{cnn} can recover characters from
synthetic \gls{lp} data that are unreadable due to strong noise and low
resolution. Their network recognizes two groups of three
characters. Lorch~\emph{et al.}~\cite{lorch2019forensic} generalize this approach to
\glspl{lp} with five to seven characters and remove constraints on the character format. 
Kaiser~\emph{et al.}~\cite{kaiser2021} evaluate Lorch~\emph{et al.}'s~\cite{lorch2019forensic} method  for synthetic and real data and report lossy compression as one of the most challenging degradation factors.
Rossi~\emph{et al.}~\cite{rossi2021denoiser} train a network that jointly consists of a U-NET denoiser and a \gls{cnn} recognition module. The method outputs both a string of \gls{lp} characters and a denoised version of the image. However, their image reconstruction requires images of slightly higher quality, and compression is not considered.
Using a \gls{crnn} method was also proposed for \gls{flpr}~\cite{moussa2021sequence}. An evaluation on synthetic data hereby showed increased recognition performance and higher robustness to out of distribution samples compared to \glspl{cnn}.

Recently, Transformer networks~\cite{vaswani2017attention} set the state of the art in natural language processing~\cite{otter2020survey} and were subsequently adapted to image processing by Dosovitskiy~\emph{et al.}~\cite{dosovitskiy2020image}.
Transformers show very promising results on tasks like image classification,
object detection or object segmentation~\cite{10.1145/3505244}.

 In this work, we demonstrate the effectiveness of Transformer architectures for \gls{flpr}. Our proposed network uses available knowledge on the input compression strength.
The integration of prior knowledge is similar to previous works on recommender
systems~\cite{hidasi2016parallel, fischer2020integrating}. Here, additional information describing items is fed to the \gls{nn} system
to improve proposals for users. To the best of our knowledge, neither
Transformers nor knowledge embedding have been investigated for \gls{flpr}.

\section{Methods}
\label{sec:methods}
We model \gls{flpr} as a \gls{seq2seq} task that operates on an image $\textbf{I} \in [0,1]^{W\times H}$ of width $W$ and height $H$. The input $\textbf{I}$ is processed column-by-column as a series of image slices $\textbf{i}_w \in [0,1]^H$, yielding the input series $\hat{\textbf{I}} = [\textbf{i}_0, \textbf{i}_1, ..., \textbf{i}_W]$. An additional input to each $\hat{\textbf{I}}$ is a scalar quantity $c_n \in [1,100]$ that encodes the JPEG compression QF of $\hat{\textbf{I}}$.
Our specific task is, given $c_n$, to translate $\hat{\textbf{I}}$ to a string output $S$ of variable length with characters $a_m \in \mathcal{A}$ of an alphabet of all valid output tokens.

In this section, we briefly summarize the concept of Transformers, present our architecture and outline libjpeg~\cite{libjpeg} QF estimation.

\subsection{Attention in Transformer Networks}
Transformers~\cite{vaswani2017attention} rely mainly on a scaled dot-product attention mechanism to grasp global dependencies between two sequences $\textbf{S}_{i}$, $\textbf{S}_{j}$. If self-attention is computed, $\textbf{S}_i = \textbf{S}_j$ are identical.
The attention function maps a matrix triple consisting of a set of queries $\textbf{Q}$, values $\textbf{V}$ and keys $\textbf{K}$ to an output. It is given by
\begin{equation}
	\textrm{Att}(\textbf{Q},\textbf{V},\textbf{K}) = \textrm{Softmax}\left(\frac{\textbf{Q}\textbf{K}^\top}{\sqrt{d_K}}\textbf{V}\right),
\end{equation}
where $d_K$ is the dimension of the queries/keys and acts as normalizing factor. $\textbf{Q}, \textbf{K}$ and $\textbf{V}$ are obtained by projecting input sequences with learnable weight matrices.  
Typically, Transformers learn multiple such projection matrices for inputs, yielding multiple $\textbf{Q}, \textbf{V}, \textbf{K}$ representations. Attention is then computed in parallel over each triple $\textbf{Q}, \textbf{V}, \textbf{K}$, which is called multi-head attention.

For solving \gls{seq2seq} tasks, a transformer network consisting of an encoder and decoder is applied. The encoder processes the input sequence and outputs it to the decoder. The decoder further takes the output sequence elements from all previous time steps as input and yields the prediction. The encoder consists of several layers, each implementing self-attention and fully-connected (FC) sub-layers. Each sub-layer includes layer-normalization as well as a residual connection around itself. The decoder is constructed similarly, but additionally computes the encoder-decoder attention, where $\textbf{K}$ and $\textbf{V}$ stem from the encoder output, while $\textbf{Q}$ stems from the decoder. For more details on Transformers, we refer to Vaswani~\emph{et al.}~\cite{vaswani2017attention}.

\subsection{Network Architecture with Knowledge Embedding}
\label{sec:know_emb}

Our \gls{seq2seq} model is shown in Fig.~\ref{fig:architecture}. First, the input $\hat{\textbf{I}}$ is combined with compression strength $c_n$ as side information via an embedding layer. The result is fed to the network consisting of a Transformer encoder (blue) and decoder (orange), each of $5$ layers with $8$ attention heads. The dimension of each (FC) sub-layer is $2160$. A final output FC layer (green) projects the decoder output in the vocabulary space 
of size $|\mathcal{A}|$, where $\mathcal{A}$ is the target vocabulary set. In our case it is dependent on the \gls{lp} format and covers $36$/$41$ tokens for Czech/German \glspl{lp} plus three special \gls{seq2seq} tokens.

The knowledge embedding with $c_n$ is computed as follows. First, $c_n$ is one-hot encoded and projected with a linear layer to the dimension of $H$. This vector is
replicated $W$ times to form $\textbf{E}_n \in \mathbb{R}^{W\times H}$.
The final embedded input sequence $\hat{\textbf{I}}_{\mathrm{emb}}$ is given by
\begin{equation}
	\hat{\textbf{I}}_{\mathrm{emb}} = \textrm{Dropout}(\hat{\textbf{I}}_{\mathrm{pos}} + \textbf{E}_n)\enspace,
	\label{eq:know}
\end{equation}
where 
$\hat{\textbf{I}}_{\mathrm{pos}}$ is the input $\hat{\textbf{I}}$ after positional encoding~\cite{vaswani2017attention}. Dropout regularization is applied with empirically chosen probability $0.5$.
\subsection{Estimating QFs}
\label{sec:estimate_qf}
To acquire a measure for compression strength, we regress the scalar JPEG QF $\in [1,100]$ of libjpeg's~\cite{libjpeg} quantization matrix ($\textbf{M}_Q$). Embedding entire matrices leads to high combinatorial complexity when densely covering the whole $\textbf{M}_Q$ space, which makes training infeasible. Following the approach of related work~\cite{cozzolino2019noiseprint}, we hence use the QF as surrogate value. In practice, the QF itself is generally not directly available from JPEG (meta)data and its value range depends on the specific $\textbf{M}_Q$ used for compression.
However, approaches for compression quality estimation exist. The $\textbf{M}_Q$ may be available from JPEG metadata or can be approximated~\cite{thai2016jpeg}. The QF can then be estimated by orthogonal projection on the libjpeg standard $\textbf{M}_Q$. To achieve this, we refer to the implementation of Cozzolino~\emph{et al.'s} method~\cite{cozzolino2019noiseprint}.

\section{Evaluation}
\label{sec:evaluation}

We train all networks for $100$ epochs and take the best run. We use the Adam optimizer and progressively reduce the initial learning rate $\eta = 1\mathrm{e}^{-4}$ by a multiplicative factor of $0.1$ with a patience of $3$ epochs. A train and validation batch size of $64$ is employed. \Gls{seq2seq} models are trained with teacher forcing~\cite{williams1989learning}. We evaluate our approach on ReId\cite{vspavnhel2017holistic} and SynthGLP and report both the accuracy per \gls{lp} (\acclp) and character error rate (CER).

\subsection{Datasets}
We use ReId~\cite{vspavnhel2017holistic} to show the general applicability of our
method towards low-quality real images. To our knowledge, there exists no
real dataset for severely degraded, unreadable \glspl{lp}. Hence, we evaluate
severe degradations in controlled experiments on synthetic data.

ReId consists of $105\:924$ train  and $76\: 412$ test samples of mostly Czech \glspl{lp} from video
cameras filming highways (Fig.~\ref{fig:reid}).  
The SynthGLP dataset is newly generated for this work.
It consists of $900\text{k}$/$100\text{k}$/$1\text{k}$ images for the
train/validation/test split. 
Each image shows a grayscale \gls{lp} in German format~\cite{german_law}, rendered at a
resolution of $180\times40$ pixels by a framework similar to Spruck~\emph{et al.'s}~\cite{spruck2021}. \gls{lp} characters are sampled from the valid alphabet per position but are otherwise random to prevent biases towards specific regions.

We compute degradations on the fly. The pipeline consists of normalization, bilinear down sampling to random pixel width $r_w \in [20,180]$ with preserved ratio, JPEG compression with random QF $\in [1, 100]$  and bilinear upsampling to the original size (Fig.~\ref{fig:jpeg_data}).

\newcommand{\iwidth}{1.8cm}
\newcommand{\jpegtab}{

	\begin{tabular}{p{0.7cm}||ccc}

		&\multicolumn{3}{c}{\small{JPEG quality factors}}\\ 
		$r_w$&  1& 20 & 50 \\  \hline
		
		\vspace{0cm}20&{\raisebox{-0.2cm}{\includegraphics[width=\iwidth]{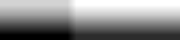}}}&\smash{\raisebox{-0.2cm}{\includegraphics[width=\iwidth]{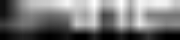}}}&
\smash{\raisebox{-0.2cm}{\includegraphics[width=\iwidth]{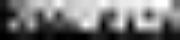}}}\\
	
		30&\includegraphics[width=\iwidth]{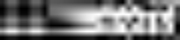}&\includegraphics[width=\iwidth]{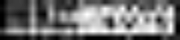} & \includegraphics[width=\iwidth]{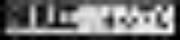}\\
		
		50&\includegraphics[width=\iwidth]{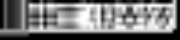}&\includegraphics[width=\iwidth]{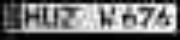} & \includegraphics[width=\iwidth]{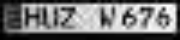}\\
		
		70&\includegraphics[width=\iwidth]{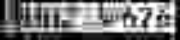}&\includegraphics[width=\iwidth]{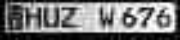} & \includegraphics[width=\iwidth]{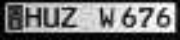}\\
		
		90&\includegraphics[width=\iwidth]{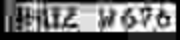}&\includegraphics[width=\iwidth]{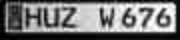} & \includegraphics[width=\iwidth]{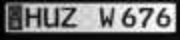}\\
	\end{tabular}

}

\begin{figure}
	\centering
	
	\subfloat[ ][ReId set~\cite{vspavnhel2017holistic} example samples]{\label{fig:reid}\includegraphics[width=0.45\textwidth]{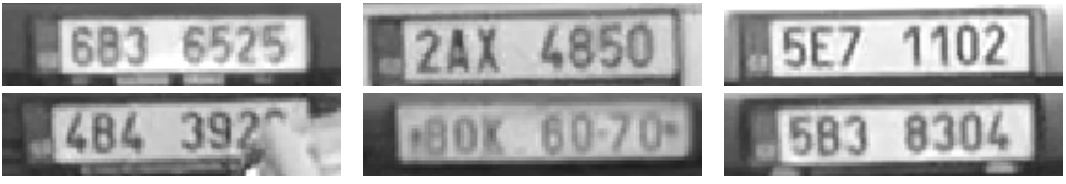}}\\
	\subfloat[][Medium and low quality selection from our SynthGLP set]{\label{fig:jpeg_data}\jpegtab}
	\caption{Samples from featured real world and synthetic sets }
\end{figure}

\subsection{Evaluation on Real World Data}
We first show that Transformers are fit for the task of real world \gls{lpr} by training our model without knowledge embedding (LP-Transf.) on the ReId~\cite{vspavnhel2017holistic} training split and evaluating on the test split.
Table~\ref{tab:reid} shows the results of running our model and related work~\cite{kaiser2021,moussa2021sequence}, as well as \v{S}pa{\v{n}}hel \emph{et al.'s}~\cite{vspavnhel2017holistic} reported values. We slightly outperform the related methods for \gls{flpr}~\cite{kaiser2021, moussa2021sequence} and share the same CER $=0.004$ as \v{S}pa{\v{n}}hel--\{S,L\}. \v{S}pa{\v{n}}hel--L achieves a slightly higher \acclp than the other two models with $ 93.6\% > 93.3\%$. All in all, we report LP-Transf. to perform comparably, while being far more parameter-efficient with $1.9$M trainable weights as opposed to $8$M and $17$M weights (\v{S}pa{\v{n}}hel--\{S,L\}).

\begin{table}[t]
	\centering
	\begin{tabular}{l|l|lll}
		Type & Method & \acclp & CER & Params \\ \hline
		\multirow{4}{*}{CNN}&\v{S}pa{\v{n}}hel-S~\cite{vspavnhel2017holistic}& $\emph{98.3\%}$    & \pmb{$0.004$}&   $\sim 8\text{M}$      \\
		&\v{S}pa{\v{n}}hel-L~\cite{vspavnhel2017holistic}&  $\pmb{98.6\%}$ &  \pmb{$0.004$}  & $\sim 17\text{M}$         \\
		&Kaiser~\cite{kaiser2021}& $97.3\%$&$0.029$       &  $\sim 45\text{M} $      \\ \hline
		CRNN &Moussa~\cite{moussa2021sequence}&    $98.1\%$&\pmb{$0.004$}&    $\sim 4.5\text{M} $      \\ \hline
		Transf.&LP-Transf.&  $\emph{98.3}\%$   & \pmb{$0.004$}&    $\pmb{\sim 1.9\text{M}} $      
	\end{tabular}
\caption{Performance on the ReId test split~\cite{vspavnhel2017holistic}. 
}
\label{tab:reid}
\end{table}

	\begin{table}[t]
		\centering
		\begin{tabular}{l|l|lll}
			Type & Method & \acclp & CER & Params \\ \hline
			\multirow{5}{*}{CNN}
			&\v{S}pa{\v{n}}hel-S~\cite{vspavnhel2017holistic}&$75.10\%$&$0.0597$	&$\sim 8\text{M}$\\
			&\v{S}pa{\v{n}}hel-L~\cite{vspavnhel2017holistic}&$79.90 \%$ &$0.0479$ &$\sim 17\text{M}$\\
			&Kaiser~\cite{kaiser2021}& $80.15\%$     & $0.0449$ &  $\sim 45\text{M} $      \\

			&EffNet-B0\cite{tan2019efficientnet}& $83.89\%$ &  $0.0378$
			&   $\sim 40\text{M}$      \\
			&EffNet-B7\cite{tan2019efficientnet}&  $87.25\%$  &$0.0304$
			& $\sim 132\text{M}$         \\
			\hline
			CRNN &Moussa~\cite{moussa2021sequence}&    $92.48\%$& $0.0221$
			& $\sim 4.5\text{M} $      \\ \hline
			\multirow{6}{*}{Transf.}&LP-Transf.&  $92.41\%$   &$0.0215$
			&  \multirow{6}{*}{ $\pmb{\sim 1.9\text{M}} $ }     \\
			&LP-Transf.-5&  $92.34\%$   &$0.0211$ \\
			&LP-Transf.-10&  $92.54\%$   &$0.0206$  \\
			&LP-Transf.-25&  $92.51\%$   &$0.0206$  \\
			&LP-Transf.-50&  $\pmb{92.83\%}$   &$\pmb{0.0195}$  \\
			&LP-Transf-100&  $\emph{92.66\%}$   &$\emph{0.0202}$ \\

		\end{tabular}
		\caption{Results on SynthGLP\textsubscript{TFull}. LP-Transf.--$\{10,25,50,100\}$ surpass all other approaches, where $K=50$ is best. LP-Transf.(--5) achieve higher CER but lower \acclp than the best baseline CRNN.}
		\label{tab:jpeg_avg}
\end{table}

\begin{figure*}[t]
	\centering
	\subfloat[][{$r_w \in [20,180]$}]{\label{fig:res_all}\includegraphics[height=4cm,valign=b]{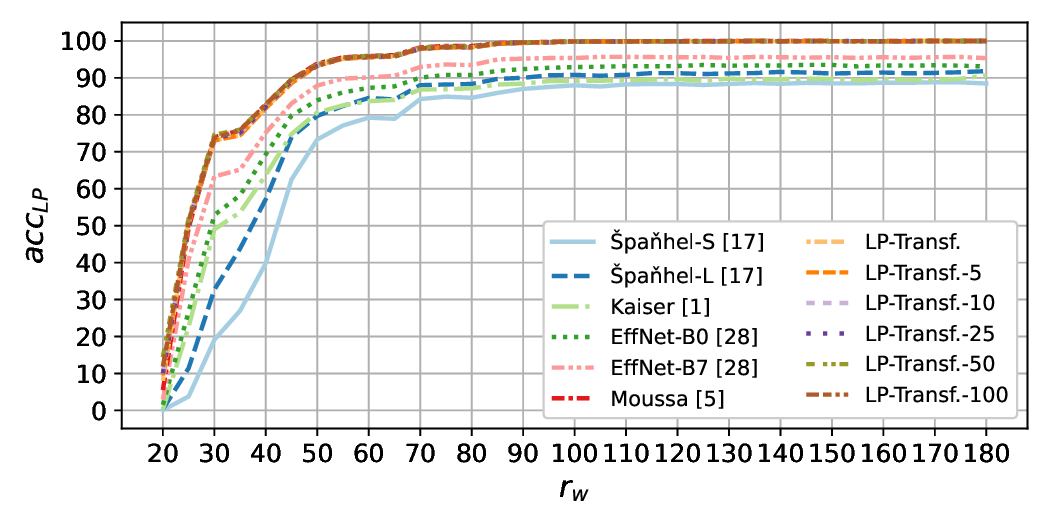}}
	\subfloat[][{$r_w \in [20,30]$}]{\label{fig:res_20-30}\includegraphics[height=4cm,valign=b]{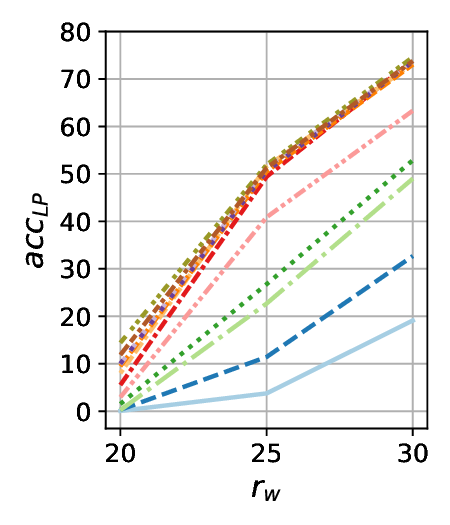}}
	\subfloat[][$r_w = 20$]{
		\adjustbox{valign=b}{	\begin{tabular}{l|l|l}
				Method &  \acclp & CER  \\ \hline
				Moussa~\cite{moussa2021sequence}&  $5.53\%$   &$0.3496$  \\
				LP-Transf.&  $8.02\%$   &$0.3254$  \\
				LP-Transf.-5 & $9.48\%$&$0.3131$ \\
				LP-Transf.-10 & $9.94\%$&$0.3046$ \\
				LP-Transf.-25 & $10.04\%$&$0.3046$ \\
				LP-Transf.-50 & $\pmb{14.43\%}$&$\pmb{0.2848}$ \\
				LP-Transf.-100&  $\emph{11.88}\%$   &$\emph{0.2990}$\\
				\vspace{0.2cm}
			\end{tabular}
			\label{tab:res_20}}}

	\label{fig:ma}

	\caption{ \acclp results averaged over $c_n \in \mathcal{C}_1$ on SynthGLP\textsubscript{TFull}.  \Gls{seq2seq} models perform comparably well for $r_w \ge 30$ and surpass all CNN methods. The advantage of knowledge embedding steadily enlarges with decreasing $r_w$ and is highest for $r_w = 20$ as depicted in Tab.~\ref{tab:res_20}. 
	}
\end{figure*}

\begin{figure*}[t]
	\includegraphics[width=\textwidth]{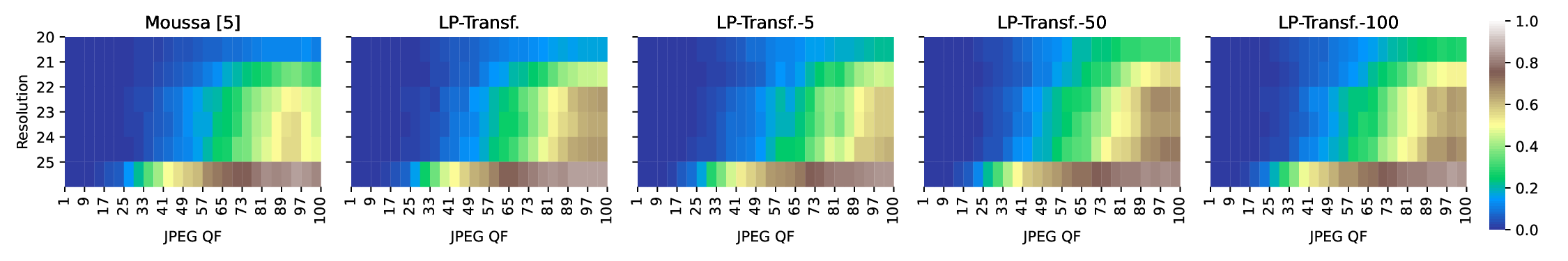}
	\caption{\acclp for the best performing baseline CRNN~\cite{moussa2021sequence} and our method. }
	\label{fig:heat_map}
\end{figure*}

\subsection{Evaluation on Synthetic Data}
We analyze the advantage of incorporating prior knowledge about compression levels on SynthGLP. For the train and validation split we use two fixed random seeds for degradation calculations to ensure identical data for all models. For testing, we generate SynthGLP\textsubscript{TFull}. It contains $858$ degraded variants of SynthGLP's test split, each featuring one of all possible $(c_n, r_w)$ combinations, with $c_n \in \mathcal{C}_1 = \{i \cdot 4 +1| i \in [0,24] \} \cup \{100\}$ and $r_w \in \mathcal{R}_1 = \{i \cdot 5 | i \in [4, 36]\}$. SynthGLP\textsubscript{TFull} is designed to evenly sample all degradation levels covered in training. To analyze performance on very low resolution images, we also generate  SynthGLP\textsubscript{TLow} covering $156$ test sets with $(c_n, r_w) \in \mathcal{C}_1 \times  \mathcal{R}_2$, $\mathcal{R}_2 = \{i | i \in [20, 25]\}$.

\subsubsection{Models}
The evaluation covers various baseline approaches, and our Transformer model with and without knowledge embedding. 

\textbf{Baseline models} include the same methods as for the experiments on ReId~\cite{kaiser2021, moussa2021sequence, vspavnhel2017holistic}. We also feature state-of-the-art EfficientNet (EffNet) models B0 and B7~\cite{tan2019efficientnet}. All models' input layers are adapted to input size $180\times40$. We extend Kaiser~\emph{et al.}'s model~\cite{kaiser2021} to $9$ output layers to account for longer possible lengths of German \glspl{lp} and pre-initialize the EffNet models with ImageNet weights~\cite{pytorchmodels} prior to adding a classification head of $9$ FC layers with size $2048$.

\textbf{LP-Transf.-$K$} is the extension of LP-Transf. by a knowledge embedding layer with $K \in \mathbb{N}$ knowledge classes (Sec.~\ref{sec:know_emb}, Fig.~\ref{fig:architecture}). $K = 100$ accounts for the full value range of the QF from libjpeg. To simulate estimated QFs, we linearly map the values to a smaller range and set $c_n = \left\lfloor  \frac{K\cdot \text{QF}}{100}\right\rfloor$ for $K \in \{50,25,10,5\}$ classes. This simulates QFs that are estimated correctly within $l \in \{2,4,10,20\}$ levels, accordingly. Hence, our hardest included scenario is covered by LP-Transf.-5, which deals with only $5$ distinguishable classes, where one class includes  $l=20$ QF levels.

 \subsubsection{Results}
 The averaged performance over SynthGLP\textsubscript{TFull} is given by Tab.~\ref{tab:jpeg_avg}.
\gls{seq2seq} methods show to solve the task better than \glspl{cnn}. The method of \v{S}panhel~\emph{et al.} did achieve less good results than on ReId (Tab.~\ref{tab:reid}), which we attribute to SynthGLP\textsubscript{TFull} containing distinctly stronger degraded data. This is also consistent with the other \gls{cnn} methods' results, where performance rises with parameter size. LP-Transf.--$\{10,25,50,100\}$ outperform all other models, where $K=50$ performs best with \acclp $=92.83\%$ and CER $=0.0195$. LP-Transf.--100 performs second best. We hypothesize that the better performance for $K=50<100$ is attributed to the nature of JPEG compression. Neighboring QFs yield similar compression quality, therefore halving the QF dimension does not lead to a significant information loss but simplifies the embedding layer's optimization.  The performance of the standard LP-Transf. model and LP-Transf.--5 is comparable to the best baseline CRNN~\cite{moussa2021sequence} model.

A further analysis showed that the advantage of incorporating estimated QF knowledge is particularly large for severely degraded samples. Fig.~\ref{fig:res_all} shows the \acclp for SynthGLP\textsubscript{TFull} averaged over $c_n \in \mathcal{C}_1$ per $r_w \in \mathcal{R}_1$, Fig.~\ref{fig:res_20-30} shows the same for $r_w \in [20,30]$. \Gls{seq2seq} approaches perform comparably down to $r_w = 30$ and surpass CNN approaches for all $r_w$ by a great margin.  
Fig.~\ref{fig:res_20-30} shows a considerable benefit of knowledge embedding for $r_w \leq 30$. Especially the performance boost for the lowest $r_w = 20$ is apparent as also indicated by the table in Fig.~\ref{tab:res_20}. While all our models outperform the CRNN~\cite{moussa2021sequence}, LP-Transf.--$50$ gains the greatest advantage with \acclp $=14.43\% > 5.53\%$ and CER $=0.2848 < 0.3496$.
This shows that incorporated QF knowledge is of increasing importance for increasing compression strength. Especially for low $r_w$, where compression removes a large amount of information from the image content, the LP-Transf.-$K$ models offer the biggest advantage.

To explore this phenomenon further, we analyze the performance on very low resolution samples via SynthGLP\textsubscript{TLow}. Fig.~\ref{fig:heat_map} shows the \acclp results for the best performing baseline CRNN model, LP-Transf. and  for the weakest $K=5$, best $K=50$ and standard $K=100$ LP-Transf.-$K$ model. Obviously, LP-Transf. already handles the very low quality input better than the CRNN. The LP-Transf.-$K$ models further increase the advantage and recognize very low quality images more robustly. For $K=50$ the performance is best, where the \acclp for the lowest $r_w = 20$ first surpasses $20\%$ for $\text{QF}=61$ and reaches $31.3\%$ for QF $=100$ while the CRNN remains constantly below $12\%$.

\section{Conclusion}
\label{sec:conclusion}
In this work, we showed the effectiveness of using compression levels of images as prior knowledge to a parameter efficient Transformer model for \gls{flpr}. We showed the applicability of Transformers for \gls{lpr} on real world data and evaluated our knowledge embedding method on SynthGLP, our dataset specifically generated to fit common image degradation challenges in forensic investigations. We match the performance of the best existing \gls{flpr} approach on high and medium quality data while needing less parameters. In addition, we strongly outperform all existing methods for low quality data with QFs only estimated correctly down to $20$ levels ($K=5$).

For future work, we plan on incorporating more quality parameters from image (meta)data in our model to drive forward the research on \gls{flpr}.

\vfill\pagebreak

\bibliographystyle{IEEEbib}
\bibliography{refs}

\end{document}